\documentclass{article}

\PassOptionsToPackage{numbers, compress}{natbib}
 \usepackage[preprint]{neurips_2026}

\usepackage[utf8]{inputenc} 
\usepackage[T1]{fontenc}    
\usepackage{hyperref}       
\usepackage{url}            
\usepackage{booktabs}       
\usepackage{amsfonts}       
\usepackage{nicefrac}       
\usepackage{microtype}      
\usepackage{xcolor}         
\usepackage{amsmath}
\usepackage{amsthm}
\newtheorem{definition}{Definition}
\usepackage{multirow}
\usepackage[table]{xcolor}
\usepackage{makecell}
\usepackage{enumitem}
\usepackage{graphicx}
\usepackage{algorithm}
\usepackage{algpseudocode}
\usepackage{subcaption}
\usepackage{wrapfig}

\title{PAPO-VLA: Planning-Aware Policy Optimization for Vision-Language-Action Models}

\author{%
  Peizheng Guo\textsuperscript{1,2}\thanks{These authors contribute equally}, Jingyao Wang\textsuperscript{1,2}\footnotemark[1], Changwen Zheng\textsuperscript{1,2}, Wenwen Qiang\textsuperscript{1,2}\thanks{Corresponding author} \\
  \textsuperscript{1}Institute of Software Chinese Academy of Sciences,\\
  \textsuperscript{2}University of Chinese Academy of Sciences\\
  \texttt{\{guopeizheng2025, wangjingyao2023, changwen, qiangwenwen\}@iscas.ac.cn} \\
}

\begin{document}

\maketitle

\begin{abstract}
Vision-Language-Action (VLA) models show promising ability in language-guided robotic tasks. However, making VLA policies reliable remains challenging, because a manipulation task is completed through closed-loop interaction, where each action affects subsequent execution. To analyze this problem, we revisit VLA policy during execution and argue that a VLA policy acts both as a planner, which makes task-oriented decisions that change the direction of execution, and as an executor, which realizes these decisions through dense continuous actions. This view suggests that improving VLA reliability requires particular attention to planning actions. Existing optimization methods can imitate actions or improve complete trajectories, but they usually do not explicitly identify planning actions or measure their importance for task success. To address this issue, we propose Planning-Aware Policy Optimization for VLA models (PAPO-VLA). PAPO-VLA first identifies planning actions by jointly considering action variation and trajectory outcome, then estimates their importance through causal sufficiency and causal necessity, and finally incorporates this importance into GRPO advantage estimation. In this way, more important planning actions receive stronger optimization emphasis, while the whole trajectory is still optimized by trajectory-level feedback. Experiments on multiple benchmarks demonstrate the effectiveness of PAPO-VLA.
\end{abstract}

\section{Introduction}
\label{sec:introduction}
Vision-Language-Action (VLA) models have shown strong potential in various language-guided robotic tasks by mapping visual observations and language instructions into executable robot actions \cite{zitkovich2023rt,team2024octo,kim2024openvla,black2024pi_0,li2025simplevla}. However, making VLA policies reliable remains challenging. A manipulation task is completed through closed-loop interaction, where each action changes the environment and affects subsequent observations and decisions \cite{li2025simplevla,ross2010efficient}. Therefore, a reliable VLA policy should not only generate locally plausible actions, but also maintain the whole execution toward task success. This raises a central question: \textbf{how to guarantee the reliability of VLA policies in robotic tasks?}

\begin{figure*}[t]
    \centering
    \includegraphics[width=\textwidth]{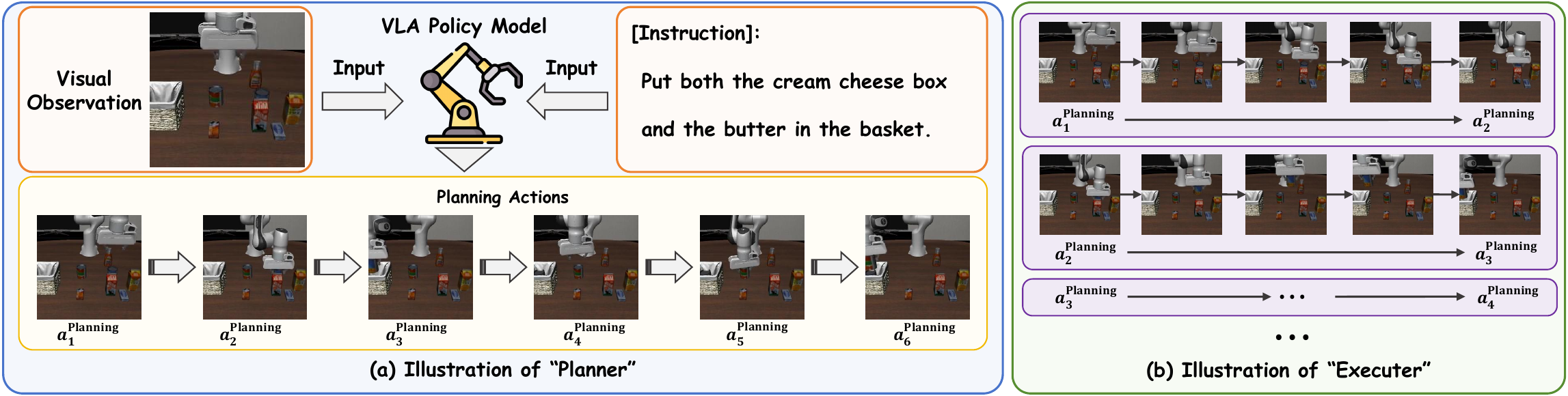}
    \caption{Illustration of Planner and Executor. (a) Planner, which consists of planning actions that are more closely related to task success; (b) Executor, which connects and realizes these planning actions through continuous motions.
    }
    \vspace{-0.15in}
    \label{fig:schema}
\end{figure*}

To answer this question, we revisit VLA policy and argue that a VLA policy plays two functional roles (\textbf{Figure \ref{fig:schema}}). First, it acts as a planner when the action changes execution intention, such as switching from approaching an object to grasping it, from grasping to moving it, or from moving to releasing it. These actions determine the direction of the following execution. Second, it acts as an executor when it produces dense actions to connect, adjust, and realize these decisions under changing observations. Both roles are necessary for successful manipulation. Nevertheless, they influence reliability in different ways. Execution actions ensure motion continuity, while planning actions determine whether the following execution is built on a correct task-oriented decision. If a planning action is unreliable, such as grasping at an unstable pose or releasing before proper alignment, later execution may become difficult to recover. Thus, improving the reliability of a VLA policy requires particular attention to the reliability of its planning actions.

This perspective leads to two key problems. The first problem is how to obtain planning actions in a dense execution trajectory. In this work, we seek a way to identify planning actions from the trajectory. Since these actions often appear when the policy changes its execution intention, action variation provides a useful cue for locating planning actions \cite{sutton1999between,konidaris2012robot}. However, action variation alone is not sufficient. An abrupt action change may correspond to a meaningful transition, but it may also result from unstable control or a failed attempt \cite{ross2011reduction,kroemer2021review}. Therefore, the planning actions should also be associated with trajectory outcomes, so that the optimization focuses more on transitions that are likely to support successful execution. The second problem is how to measure the importance of these planning actions for task success. A planning action is more important if preserving it helps the subsequent execution move toward success, and if perturbing it makes the task outcome worse. Thus, we need to identify planning actions and estimate their importance for task success.

Based on this analysis, we propose Planning-Aware Policy Optimization for VLA models (PAPO-VLA). Our method first identifies planning actions by action variation and trajectory outcome. This allows policy optimization to focus on actions that indicate execution transitions and appear in trajectories with better outcomes. Then, we estimate the importance of each planning action through causal sufficiency and necessity \cite{pearl2009causality}. Causal sufficiency measures whether preserving the action helps the subsequent execution reach a successful outcome. Causal necessity measures whether perturbing the action makes the outcome worse. An action receives high importance when it is both sufficient for supporting successful progress and necessary for preserving task success. Finally, we incorporate this importance into GRPO advantage estimation, so that more important planning actions receive stronger emphasis while other actions are still optimized by the trajectory outcome.

The main contributions of this paper are summarized as follows:
(i) We revisit VLA policy and argue that it plays two functional roles during manipulation: a planner role that determines task-oriented execution decisions, and an executor role that realizes these decisions through dense actions. This view highlights that the reliability of planner is crucial for the reliability of VLA policy.
(ii) We propose Planning-Aware Policy Optimization for VLA models (PAPO-VLA), which identifies planning actions, estimates their importance through causal sufficiency and necessity, and incorporates the resulting importance into GRPO advantage estimation;
(iii) Extensive experiments conducted on various benchmarks indicate the effectiveness of our method.

\section{Related Work}
\label{sec:related_work}

\textbf{Vision-Language-Action Model.}
Vision-Language-Action (VLA) models aim to enable robots to generate executable actions in complex environments. Recent studies mainly focus on scaling robotic policies with large pretraining, and exploring post-training to adapt pretrained policies to downstream tasks. OpenVLA \cite{kim2024openvla} unifies open-source vision-language modeling with robot control. $\pi_{0.7}$ \cite{intelligence2026pi} uses richer multimodal prompts to support manipulation across tasks, robots, and scenes. RIPT-VLA \cite{tan2025interactive} shows that post-training with task-success rewards can effectively improve the adaptability of pretrained VLA policies in new settings. TGRPO \cite{chen2025tgrpo} introduces trajectory-wise group relative policy optimization into VLA fine-tuning, providing a direct exploration of GRPO-style post-training for VLA models.
Unlike previous studies, we argue that VLA policy plays two roles during execution: planner and executor. We focus on the reliability of planner to guarantee VLA policy reliability by identify planning actions from dense trajectories, estimate their importance for task success, and incorporate this importance into GRPO-based optimization. 

\textbf{Causality in Robot Learning.}
Causal theory provides a theoretical framework for modeling interventions, counterfactual changes, and task invariance in robot learning, and has recently been used to improve policy performance. CAIAC \cite{urpi2024causal} identifies state components that are not affected by actions to improve the robustness of offline robot learning. RoCoDA \cite{ameperosa2025rocoda} combines causal invariance, geometric equivariance, and data augmentation to improve imitation learning under changing settings. LaVA-Man \cite{zhu2025lava} uses goal-image prediction as a pretraining task to implicitly learn causal relations between visual states and actions in language-guided manipulation.
These efforts show that causality can provide more structured modeling principles for robot policy learning, but mainly focus on state factors, data augmentation, or representation learning. 
We use causality to measure the importance of each planning action during policy optimization. By measuring causal sufficiency and necessity of such planning actions, and incorporating them into optimization, our method enables the optimization to more explicitly distinguish actions that are important for ensuring task success.

\section{Problem Setting and Analysis}
\label{sec:problem_setting_analysis}
In this section, we first introduce the problem setting of VLA in \textbf{Section \ref{sec:problem_setting}}. Then, we provide the motivation and causal analysis of VLA in \textbf{Section \ref{sec:motivation}}. More discussion is provided in \textbf{Appendix}.

\subsection{Problem Setting}
\label{sec:problem_setting}
In this paper, we consider language-conditioned robotic manipulation, where a VLA receives visual observations and a task instruction, and generates an action sequence to complete the task. Specifically, let $l$ denote the language instruction and $o_t$ denote the visual observation at step $t$. At each step, the policy takes the most recent $H$ observations and the instruction as input, and outputs an action $a_t$:
\begin{equation}
    a_t \sim \pi_\theta(a_t \mid o_{t-H+1:t}, l),
\end{equation}
where $o_{t-H+1:t}=\{o_{t-H+1},\dots,o_t\}$ denotes the observation window up to time step $t$, and $\theta$ denotes the policy parameters. This formulation covers both the case where the policy depends only on the current observation and the case where it conditions on a short history of observations.

The interaction between the policy and the environment forms a trajectory $\tau=(o_0,a_0,o_1,a_1,\dots,o_{T_i-1},a_{T_i-1})$, where $T_i$ denotes the trajectory length. The corresponding action sequence is $\mathcal{A}=\{a_0,a_1,\dots,a_{T_i-1}\}$.

Following GRPO \cite{guo2025deepseek}, we optimize the VLA policy by comparing multiple candidate trajectories generated under the same task condition. Specifically, given an initial input $x=(o_0,l)$, the old policy $\pi_{\theta_{\mathrm{old}}}$ samples a group of $G$ candidate trajectories $\{\tau^i\}_{i=1}^{G}$. Each trajectory $\tau^i$ is associated with a scalar trajectory-level reward $r^i$, which can be computed from task success, trajectory return, or their combination \cite{guo2025deepseek}. We optimize the policy by maximizing the following GRPO objective:
\begin{equation}
\begin{aligned}
    J_{\mathrm{GRPO}}(\theta)&=
\mathbb{E}_{x\sim P,{\tau^i}\sim\pi*{\theta_{\mathrm{old}}}}\\
\Big[
\frac{1}{G}\sum_{i=1}^{G}\frac{1}{T_i}\sum_{t=0}^{T_i-1}
\Big(
\min\big(\rho_{i,t}(\theta)A_i,&
\mathrm{clip}(\rho_{i,t}(\theta),1-\epsilon,1+\epsilon)A_i\big)
-\beta D_{\mathrm{KL}}(\pi_\theta|\pi_{\mathrm{ref}})
\Big)
\Big],
\end{aligned}
\end{equation}
where $P$ denotes the input distribution, $\mathrm{clip}(\rho_{i,t}(\theta),1-\epsilon,1+\epsilon)$ constrains the update magnitude with clipping threshold $\epsilon$, and $D_{\mathrm{KL}}(\pi_\theta|\pi_{\mathrm{ref}})$ denotes the KL divergence between the current policy $\pi_\theta$ and a fixed reference policy $\pi_{\rm ref}$. The action-level importance ratio is defined as
\begin{equation}
    \rho_{i,t}(\theta)=\frac{\pi_\theta(a_t^i\mid o_{t-H+1:t}^i,l)}{\pi_{\theta_{\mathrm{old}}}(a_t^i\mid o_{t-H+1:t}^i,l)}.
\end{equation}

The relative advantage $A_i$ measures the relative quality of each trajectory within the group:
\begin{equation}
    A_i=\frac{r^i-\mathrm{mean}(r^1,\dots,r^G)}{\mathrm{std}(r^1,\dots,r^G)},
\end{equation}
where $\mathrm{mean}(r^1,\dots,r^G)$ and $\mathrm{std}(r^1,\dots,r^G)$ denote the mean and standard deviation of trajectory rewards within the same group, respectively.

\begin{figure*}[t]
    \centering
    \includegraphics[width=\textwidth]{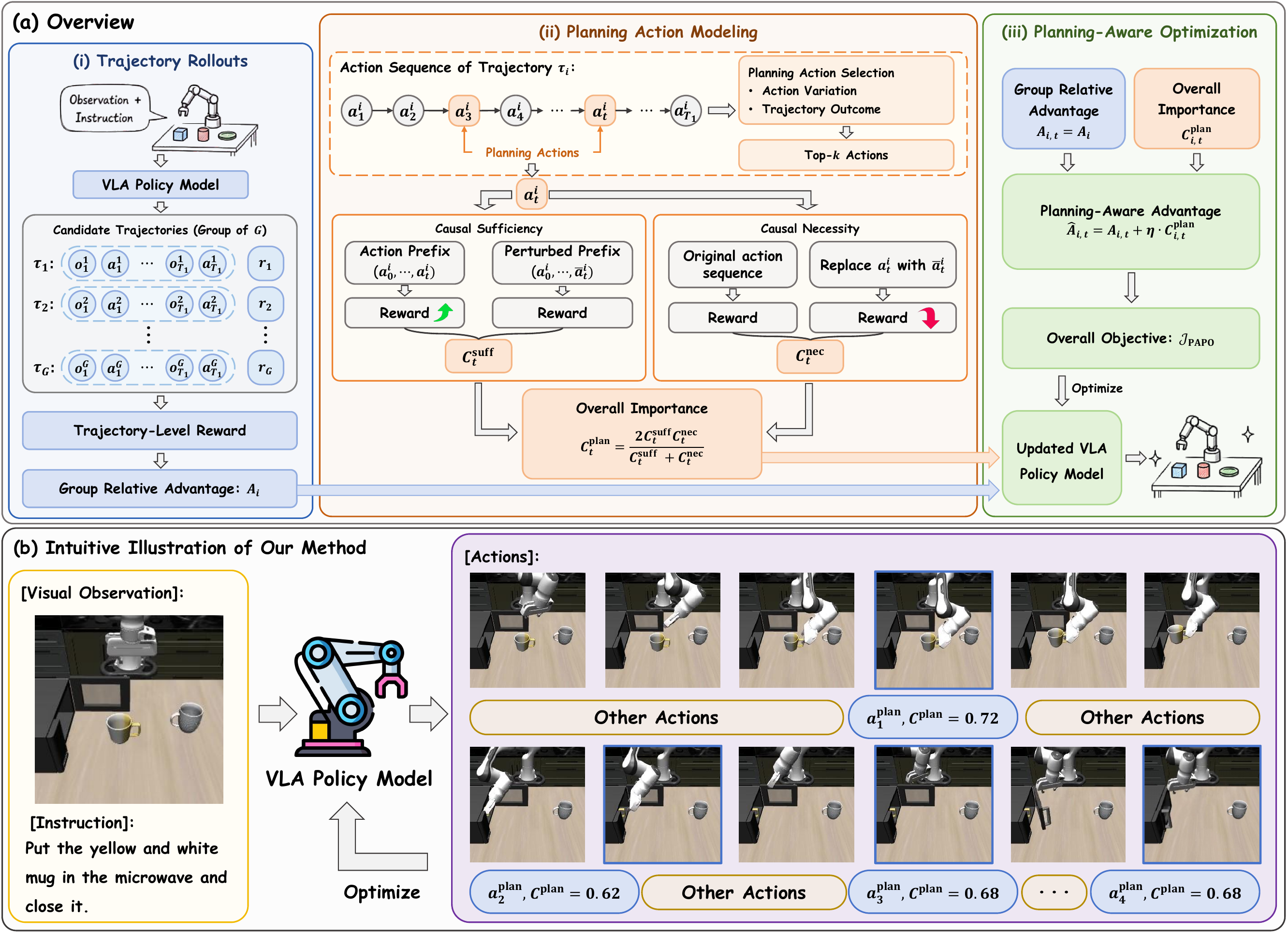}
    \caption{Overview of our proposed method. (a) Overview: Trajectory rollouts are first evaluated by trajectory-level rewards, and planning actions are selected and analyzed through causal sufficiency and necessity to obtain their importance, incorporated into the GRPO advantage estimation to guide optimization. (b) Intuitive Illustration: how different actions in a manipulation trajectory receive different importance to optimize the VLA policy.
    }
    \vspace{-0.2in}
    \label{fig:overview}
\end{figure*}

\subsection{Motivation Analysis}
\label{sec:motivation}
A VLA policy completes a manipulation task through a closed-loop action sequence and each action changes the environment and further affects later observations and decisions \cite{brohan2022rt,zitkovich2023rt,kim2024openvla,black2024pi_0,team2025gemini}. Therefore, the reliability of a VLA policy should not be understood only from whether each action is locally plausible, but from whether the whole execution can be maintained toward task success.

To analyze this problem, we revisit VLA policy during execution. Although the policy is usually formulated as a continuous action generator, it plays two functional roles in a trajectory (\textbf{Figure \ref{fig:schema}}). When the actions change execution intention, such as switching from approaching to grasping, from grasping to moving, or from moving to releasing, the policy acts as a planner. We refer to these task-oriented decision steps as planning actions. When the policy generates dense motions to connect, adjust, and realize these decisions, it acts as an executor. Both roles are necessary, but planning actions have a stronger influence on whether the following execution stays on a successful path.

This makes planning actions important for VLA reliability. If a planning action is unreliable, later execution may become ineffective even if the following actions are smooth. For example in \textbf{Figure \ref{fig:schema}}, an unstable grasp can make subsequent moving actions useless, and an early release can make the remaining trajectory difficult to recover. However, existing optimization methods usually do not explicitly distinguish such actions. Imitation learning reproduces demonstration actions, but does not indicate which actions are more important for task success. GRPO-style optimization uses trajectory-level rewards, but the same trajectory-level advantage may be assigned to all actions in a rollout, making planning actions and ordinary execution actions receive similar emphasis.

This motivates two problems. First, how can we identify planning actions from a dense execution trajectory? Since planning actions often appear when the policy changes its execution intention, action variation provides a useful cue \cite{sutton1999between,konidaris2012robot,niekum2015learning,tanneberg2021skid}. However, action variation alone is not sufficient, because abrupt changes may also come from unstable control or failed attempts \cite{ross2011reduction,kroemer2021review}. Thus, we further consider trajectory outcomes when identifying planning actions. Second, how can we measure the importance of these planning actions for task success? A planning action should be more important if preserving it helps the subsequent execution move toward success, and if perturbing it makes the task outcome worse. Therefore, we use causal sufficiency and necessity to estimate the importance of planning actions and guide policy optimization. Here, we provide the concept of probabilistic causal importance.

\begin{definition}[Probabilistic Causal Importance]\label{def:causal_vla}
Let $X=(O,L)$ denote the policy input, where $O$ is the observation and $L$ is the language instruction, and let $\mathcal{A}=\{a_1,a_2,\dots,a_T\}$ denote the action sequence generated by the VLA policy. Let $E(\mathcal{A},X)\in{0,1}$ denote the task-completion event under input $X$, where $E(\mathcal{A},X)=1$ means successful task success and $E(\mathcal{A},X)=0$ otherwise. For an action $a_t$, $\bar {\mathcal{A}}_t=(a_1,\dots,\bar a_t,\dots,a_T)$ denotes the perturbed action sequence by replacing the $t^{th}$ action with a perturbed action $\bar a_t$. Then, the probability that $a_t$ satisfies causal sufficiency and necessity for task success is defined as:
\begin{equation}
    C(a_t)=P_{\mathrm{suff}}P\left(E(\bar {\mathcal{A}}_t,X)=0,\bar {\mathcal{A}}_t,X\right)+P_{\mathrm{nec}}P\left(E(\mathcal{A},X)=1,\mathcal{A},X\right),
\end{equation}
where $P_{\mathrm{suff}}=P\left(E_{do(a_t)}(\bar {\mathcal{A}}_t,X)=1\mid E(\bar {\mathcal{A}}_t,X)=0,\bar {\mathcal{A}}_t,X\right)$ denotes causal sufficiency, i.e., the probability that the task becomes successful when we intervene with $do(a_t)$ at step $t$, given that the perturbed sequence $\bar{\mathcal{A}}_t$ would fail. Similarly, $P_{\mathrm{nec}}=P\left(E_{do(\bar a_t)}(\mathcal{A},X)=0\mid E(\mathcal{A},X)=1,\mathcal{A},X\right)$ denotes causal necessity, i.e., the probability that the task becomes unsuccessful when we intervene with $do(\bar a_t)$ at step $t$, given that the sequence $\mathcal{A}$ succeeds.
\end{definition}
\textbf{Definition \ref{def:causal_vla}} characterizes the importance of action $a_t$ in task success. $P_{\mathrm{suff}}$ measures whether keeping or enforcing this action makes a previously unsuccessful execution more likely to succeed. $P_{\mathrm{nec}}$ measures whether perturbing or replacing this action makes a previously successful execution more likely to deteriorate. These two terms correspond to two intuitive questions: whether the action can support success, and whether the action is important for preserving success. Therefore, when an action has both high sufficiency and high necessity, it indicates that the action not only helps drive the task toward successful completion, but also has a substantial influence on the final outcome once it is changed. Such an action should be regarded as more critical to task success, and therefore should receive stronger emphasis in optimization.

\section{Methods}
\label{sec:method}
Based on the above analysis, we propose Planning-Aware Policy Optimization for VLA models (PAPO-VLA). The core idea is to identify planning actions in trajectories, estimate their importance for task success to guide policy optimization. 
In \textbf{Section \ref{sec:action_causal_contribution}}, we first identify planning actions according to action variation and trajectory outcome. We then measure the importance of these planning actions from causal sufficiency and causal necessity, and aggregate them into an overall planning-action importance. 
In \textbf{Section \ref{sec:cgpo_vla}}, incorporate the planning-action importance into GRPO advantage estimation, so that actions that are more important for task success receive stronger optimization emphasis. The overview is shown in \textbf{Figure \ref{fig:overview}}, with the pseudo-code in \textbf{Appendix}.

\subsection{Planning Action Modeling}
\label{sec:action_causal_contribution}
In this section, we model planning actions in a VLA trajectory. First, we identify planning actions according to action variation and task outcome (\textbf{Section \ref{sec:action_selection}}). Then, we evaluate their importance from two causal perspectives. Causal sufficiency measures whether preserving a planning action better supports subsequent successful execution, while causal necessity measures whether perturbing the planning action causes the task outcome to degrade (\textbf{Section \ref{sec:causal_sufficiency}\&\ref{sec:causal_necessity}}). Finally, we combine these two quantities into an overall planning-action importance (\textbf{Section \ref{sec:causal_overall}}).

\subsubsection{Planning Action Identification}
\label{sec:action_selection}
Let a trajectory be written as $\tau=(o_0,a_0,o_1,a_1,\dots,o_{T-1},a_{T-1})$ with the corresponding action sequence $\mathcal{A}=\{a_0,a_1,\dots,a_{T-1}\}$. 
Planning actions are task-oriented decision steps in the trajectory. They often appear when the policy changes its execution intention, such as switching from approaching to grasping or from moving to releasing. Such actions usually show larger changes compared with neighboring actions. At the same time, a large action change does not necessarily indicate a planning action, because it may also come from unstable control or a failed attempt. Therefore, we identify planning actions by jointly considering action variation and trajectory outcome.

We first compute the variation magnitude of the $t^{th}$ action:
\begin{equation}\label{eq:candidate}
    \scalebox{0.85}{$u_t=
    \left\{\begin{matrix}
    \dfrac{1}{d}\left \| a_0 \right \|_1, &t=0,\\
    \dfrac{1}{d}\left \| a_t-a_{t-1} \right \|_1, &t>0,
    \end{matrix}\right.$}
\end{equation}
where $d$ is the action dimension, and $|\cdot|_1$ denotes the sum of absolute values over all action dimensions, so \textbf{Eq. \ref{eq:candidate}} can be understood as the average magnitude of difference between the current action and the previous one across dimensions. A larger $u_t$ indicates a more noticeable action change. To reduce the influence of action scale across trajectories, we normalize it as $\tilde u_t=\frac{u_t}{\frac{1}{T}\sum_{j=0}^{T-1}u_j}$.

Then, we introduce a outcome-aware gate derived from the trajectory outcome. Let $g(\tau)=\frac{r(\tau)-r_{\rm min}}{r_{\rm max}-r_{\rm min}}\in[0,1]$, where $r_{\rm min}$ and $r_{\rm max}$ denote the lower and upper bounds of the trajectory reward $r(\tau)$. A larger $g(\tau)$ indicates a better task outcome. Then, the planning-action score is defined as
\begin{equation}
    s_t=\tilde u_tg(\tau).
\end{equation}
We then select the top-$k$ actions according to this score and obtain the planning-action mask $m^{\rm plan}_t=\mathrm{TopKMask}(s_t,k)$. 
The planning actions of trajectory $\tau$ is then represented by the selected action index set $\mathcal{K}_{\tau}=\{t\mid m_t^{\mathrm{plan}}=1\}$. This identification prioritizes actions that both show noticeable execution changes and appear in trajectories with better outcomes.

\subsubsection{Causal Sufficiency Importance}
\label{sec:causal_sufficiency}
According to Definition \ref{def:causal_vla}, causal sufficiency concerns whether keeping this action can better support the generation of a correct subsequent action sequence and eventually task success. For a planning action $a_t$, 
we define the causal sufficiency importance of action $a_t$ as
\begin{equation}\label{eq:suff}
    C_t^{\mathrm{suff}}=m^{\rm plan}_t\cdot\left[\mathbb{E}_{\tau\sim \pi(\cdot|\mathcal{A}_{\le t})}r(\tau)-\mathbb{E}_{\tau\sim \pi(\cdot|\bar{\mathcal{A}}_{\le t})}r(\tau)\right]_+,
\end{equation}
where $[\cdot]_+=\max (\cdot,0)$, $r(\tau)$ denote the trajectory-level outcome reward, $\mathcal{A}_{\le t}$ denote the action prefix keeping the candidate action $a_t$, and $\bar{\mathcal{A}}_{\le t}=(a_0,\cdots,\bar a_t)$ denote the perturbed prefix replacing $a_{t}$ with a feasible small perturbation $\bar a_t$. \textbf{Eq. \ref{eq:suff}} reflects sufficiency for directly comparing the trajectory outcomes of two cases: keeping the current action versus replacing it with a perturbation. If keeping $a_t$ leads to a higher expected reward for the subsequent trajectory, $a_t$ provides more sufficient support for subsequent action generation and positively contributes to task success. Therefore, a larger $C_t^{\mathrm{suff}}$ indicates stronger causal sufficiency importance of $a_t$.

\subsubsection{Causal Necessity Importance}
\label{sec:causal_necessity}
Unlike causal sufficiency, causal necessity concerns whether the task outcome deteriorates significantly once the current action is replaced by another feasible action, given that the original action sequence can already support task success. To this end, for a planning action $a_t$, the causal necessity importance of action $a_t$ can be defined as
\begin{equation}\label{eq:nec}
    C_t^{\mathrm{nec}}=m^{\rm plan}_t\cdot\left[\mathbb{E}_{\tau\sim\pi(\cdot\mid \mathcal{A})}r(\tau)-\mathbb{E}_{\tau\sim\pi(\cdot\mid \bar{\mathcal A}_t)} r(\tau)\right]_+.
\end{equation}
where $[\cdot]_+=\max (\cdot,0)$, $r(\tau)$ denote the trajectory-level outcome reward, and $\bar{\mathcal A}_t=(a_0,\dots,\bar a_t,\dots,a_{T-1})$ denote the action sequence obtained by replacing the action $a_t$ with a feasible small perturbation $\bar a_t$. \textbf{Eq. \ref{eq:nec}} reflects necessity for directly measuring the expected difference in trajectory outcomes before and after replacing the planning action. If replacing $a_t$ with $\bar a_t$ causes a clear drop in the expected reward of the subsequent trajectory, $a_t$ is important for maintaining the task success. Therefore, a larger $C_t^{\mathrm{nec}}$ indicates stronger causal necessity importance of $a_t$.

\subsubsection{Overall Planning-Action Importance}
\label{sec:causal_overall}
After obtaining the causal sufficiency importance $C_t^{\mathrm{suff}}$ and the causal necessity importance $C_t^{\mathrm{nec}}$ of action $a_t$, we further aggregate them into a unified quantity to characterize its overall causal importance to task success. Intuitively, an actual critical action should not only provide sufficient support for successful future behavior, but also cause noticeable degradation once it is replaced. Therefore, a planning action should be regarded as having strong importance when both sufficiency and necessity are simultaneously high. So we define the overall importance of action $a_t$ as
\begin{equation}\label{eq:causal_overall}
    C_t^{\mathrm{plan}}=\frac{2C_t^{\mathrm{suff}}C_t^{\mathrm{nec}}}{C_t^{\mathrm{suff}}+C_t^{\mathrm{nec}}}.
\end{equation}
\textbf{Eq. \ref{eq:causal_overall}} provides a characterization of the requirement of being both sufficient and necessary. If an action is high in both sufficiency and necessity, then $C_t^{\mathrm{plan}}$ will also be large. In contrast, if the action is high in only one aspect but weak in the other, its $C_t^{\mathrm{plan}}$ will be suppressed. Thus, we use $C_t^{\mathrm{plan}}$ as the unified planning-action importance for the policy optimization and the causal importance of non-planning positions is explicitly set to zero.

\subsection{Planning-Aware Policy Optimization for VLA}
\label{sec:cgpo_vla}
In this section, we provide a feasible approach that incorporates planning-action importance into the VLA policy optimization. GRPO is a representative framework for RL, especially in settings where policy quality can be evaluated by verifiable trajectory outcomes \cite{guo2025deepseek}. It uses group-wise relative comparison to assess candidate trajectories, which provides a stable optimization signal without requiring additional value modeling. Therefore, we integrate the planning-action importance on the basis of GRPO for VLA optimization, constructing PAPO-VLA.

\subsubsection{Planning-Aware Advantage Estimation}
\label{sec:causal_advantage}
In \textbf{Section \ref{sec:problem_setting}}, GRPO assigns each sampled trajectory $\tau^i$ a group-relative advantage $A_i$, which reflects the overall quality of the trajectory. $A_i$ treats all actions within the same trajectory equally, i.e. $A_{i,t}=A_i$. We incorporate the planning-action importance obtained in \textbf{Section \ref{sec:action_causal_contribution}} into the advantage. Specifically, for action $a_t^i$ in trajectory $\tau^i$, let $C_{i,t}^{\mathrm{plan}}$ denote its overall planning-action importance. We then define the planning-aware advantage as
\begin{equation}\label{eq:causal_advantage}
    \widehat A_{i,t}=A_{i,t}+\eta\cdot C_{i,t}^{\mathrm{plan}},
\end{equation}
where $\eta$ controls the strength of planning-action importance. \textbf{Eq. \ref{eq:causal_advantage}} is no longer determined only by whether a trajectory is globally good or bad, but also by planning actions.

\begin{figure*}[t]
    \centering
    \includegraphics[width=\textwidth]{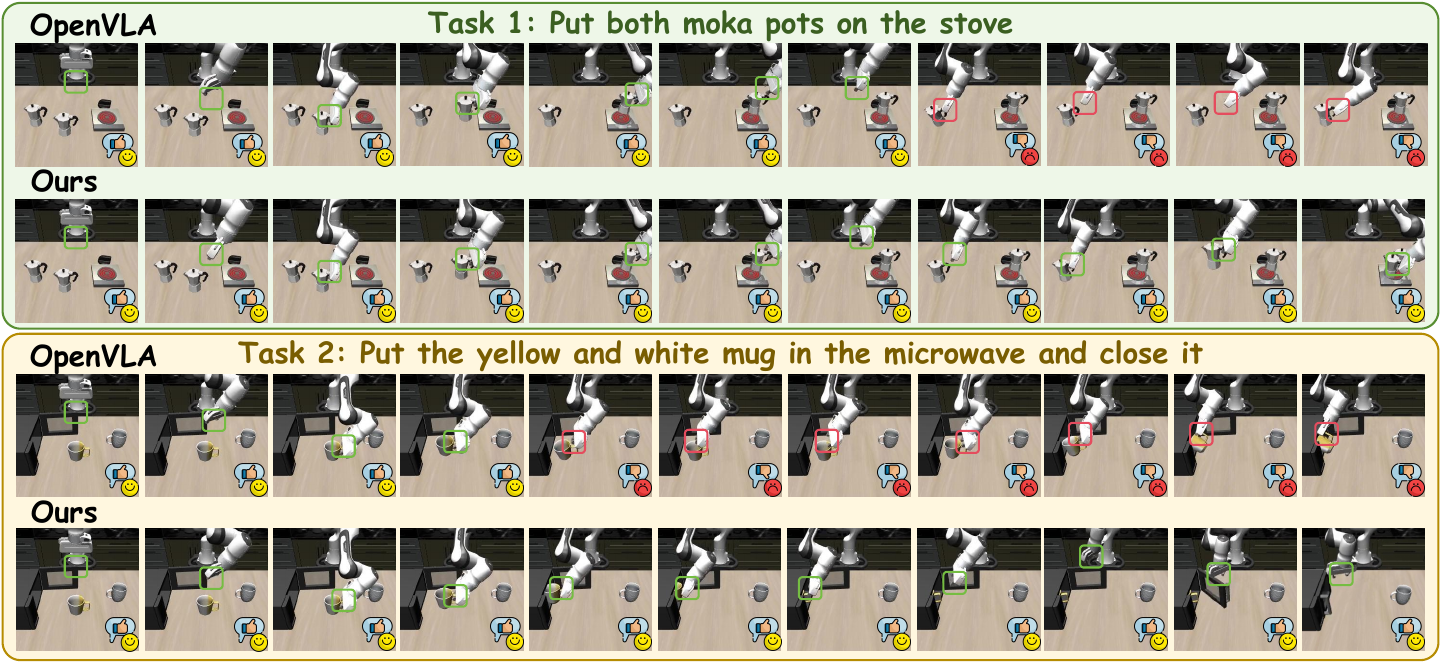}
    \caption{Visualization of comparison between OpenVLA and Ours.
    }
    \vspace{-0.1in}
    \label{fig:case}
\end{figure*}

\subsubsection{Planning-Aware Policy Optimization}
Based on the advantage in \textbf{Section \ref{sec:causal_advantage}}, we construct the optimization objective under the GRPO framework. 
Following GRPO \cite{guo2025deepseek}, the overall objective is
\begin{equation}\label{eq:causal_grpo}
\begin{aligned}
    J_{\mathrm{PAPO}}(\theta)&=
\mathbb{E}_{x\sim P,{\tau^i}\sim\pi*{\theta_{\mathrm{old}}}}\\
\Big[
\frac{1}{G}\sum_{i=1}^{G}\frac{1}{T_i}\sum_{t=0}^{T_i-1}
\Big(
\min\big(\rho_{i,t}(\theta)\widehat A_{i,t},&
\mathrm{clip}(\rho_{i,t}(\theta),1-\epsilon,1+\epsilon)\widehat A_{i,t}\big)
-\beta D_{\mathrm{KL}}(\pi_\theta|\pi_{\mathrm{ref}})
\Big)
\Big],
\end{aligned}
\end{equation}
where $\rho_{i,t}(\theta)$, $\epsilon$, and $\beta$ follow the same definitions as in \textbf{Section \ref{sec:problem_setting}}. \textbf{Eq. \ref{eq:causal_grpo}} retains the strengths of GRPO while explicitly incorporating planning-action importance into policy learning. Actions that are both more sufficient and necessary will exert a stronger effect. The optimization no longer depends only on whether a trajectory is successful, but also focuses on planning actions that are more important for task success.

\section{Experiments}
\label{sec:experiment}
In this section, we conduct extensive experiments on various benchmarks to verify the effectiveness of our method. More details, experiments, and analysis are provided in \textbf{Appendix}.

\subsection{Experimental Setup}
\label{sec:exp_setup}
\begin{wraptable}{r}{0.55\textwidth}
\vspace{-1.0em}
\centering
\caption{Main results on the LIBERO benchmark.}
\label{tab:libero_main}
\setlength{\tabcolsep}{2.5pt}
\renewcommand{\arraystretch}{0.92}
\small
\begin{tabular}{lccccc}
\toprule
\textbf{Model} & \textbf{Spatial} & \textbf{Object} & \textbf{Goal} & \textbf{Long} & \textbf{Avg.} \\
\midrule
PackNet \cite{mallya2018packnet} & 0.63 & 0.60 & 0.75 & 0.25 & 0.56 \\
MTL \cite{liu2023libero} & 0.83 & 0.54 & 0.80 & 0.48 & 0.66 \\
ATM \cite{wen2023any} & 0.69 & 0.68 & 0.78 & 0.39 & 0.63 \\
Octo \cite{team2024octo} & 0.78 & 0.85 & 0.84 & 0.51 & 0.75 \\
OpenVLA \cite{kim2024openvla} & 0.85 & 0.88 & 0.79 & 0.53 & 0.76 \\
OpenVLA-OFT \cite{kim2024openvla} & 0.91 & 0.95 & 0.90 & 0.86 & 0.91 \\
TraceVLA \cite{zheng2024tracevla} & 0.85 & 0.85 & 0.75 & 0.54 & 0.75 \\
GRAPE \cite{zhang2024grape} & 0.88 & 0.92 & 0.83 & 0.57 & 0.80 \\
SFT-4LIBERO \cite{li2025metavla} & 0.85 & 0.87 & 0.77 & 0.55 & 0.76 \\
MetaVLA \cite{li2025metavla} & 0.88 & 0.88 & 0.79 & 0.55 & 0.78 \\
TGRPO \cite{chen2025tgrpo} & 0.90 & 0.92 & 0.81 & 0.59 & 0.81 \\
Nora \cite{hung2025nora} & 0.92 & 0.95 & 0.89 & 0.74 & 0.87 \\
\midrule
\rowcolor{green!5}
\textbf{Ours} & \textbf{0.93} & \textbf{0.98} & \textbf{0.98} & \textbf{0.94} & \textbf{0.96} \\
\bottomrule
\end{tabular}
\vspace{-1.0em}
\end{wraptable}

\textbf{Benchmark.}
We evaluate our method on two benchmarks, LIBERO \cite{liu2023libero} and RoboTwin2.0 \cite{chen2025robotwin}, to examine its effectiveness on language-guided robotic manipulation tasks. All experiments are conducted on H100 GPU clusters. We report task success rate as the main evaluation metric.

LIBERO \cite{liu2023libero} contains four 10-task suites in the LIBERO Franka Panda simulation environment, including LIBERO-Spatial, LIBERO-Object, LIBERO-Goal, and LIBERO-Long. Each task provides RGB observations, robot states, task instructions, and delta end-effector actions. For evaluation, each task is tested with 50 held-out rollouts. The policy is trained for 50 episodes before evaluation.

RoboTwin2.0 \cite{chen2025robotwin} contains 50 tasks with 731 object instances and introduces comprehensive domain randomization, including variations in clutter, lighting, background, tabletop height, and language instructions. In our experiments, we use the Agilex Piper robotic arm under domain-randomized task settings. Each task is evaluated on 100 held-out test scenarios. We select 12 tasks from RoboTwin2.0 and categorize them into four horizon levels according to their average step counts.

\textbf{Baseline.}
We compare with several representative baselines: PackNet \cite{mallya2018packnet}, MTL \cite{liu2023libero}, ATM \cite{wen2023any}, Octo \cite{team2024octo}, OpenVLA \cite{kim2024openvla}, OpenVLA-OFT \cite{li2025simplevla}, RDT \cite{liu2024rdt}, TraceVLA \cite{zheng2024tracevla}, GRAPE \cite{zhang2024grape}, $\pi_{\rm fast}$ \cite{pertsch2025fast}, Nora \cite{hung2025nora}, OpenVLA-OFT \cite{li2025simplevla}, SFT-4LIBERO \cite{li2025metavla}, MetaVLA \cite{li2025metavla}, and TGRPO \cite{chen2025tgrpo}.

\begin{table*}[t]
\centering
\caption{Main results of different VLA models on RoboTwin2.0, organized by task horizon.}
\label{tab:robotwin_main_horizon}
\resizebox{\linewidth}{!}{
\begin{tabular}{lccccc}
\toprule
\multirow{2}{*}{\textbf{Model}} & \multicolumn{5}{c}{\textbf{Short Horizon Tasks (100--130 Steps)}} \\
\cmidrule(lr){2-6}
& \textbf{Lift Pot} & \textbf{Beat Hammer Block} & \textbf{Pick Dual Bottles} & \textbf{Place Phone Stand} & \textbf{Avg.} \\
\midrule
$\pi_{0}$ \cite{black2024pi_0} & 51.0 & 59.0 & 50.0 & 22.0 & 45.5 \\
RDT \cite{liu2024rdt} & 45.0 & 22.0 & 18.0 & 13.0 & 24.5 \\
OpenVLA-OFT \cite{li2025simplevla} & 10.1 & 28.1 & 29.7 & 17.1 & 21.3 \\
\midrule
\rowcolor{green!5}
\textbf{Ours} & \textbf{62.7} & \textbf{78.5} & \textbf{60.2} & \textbf{33.1} & \textbf{58.6} \\
\midrule

\multirow{2}{*}{\textbf{Model}} & \multicolumn{5}{c}{\textbf{Medium Horizon Tasks (150--230 Steps)}} \\
\cmidrule(lr){2-6}
& \textbf{Move Can Pot} & \textbf{Place A2B Left} & \textbf{Place Empty Cup} & \textbf{Handover Mic} & \textbf{Avg.} \\
\midrule
$\pi_{0}$ \cite{black2024pi_0} & 41.0 & 38.0 & 84.0 & \textbf{96.0} & 64.8 \\
RDT \cite{liu2024rdt} & 33.0 & 21.0 & 42.0 & 95.0 & 47.8 \\
OpenVLA-OFT \cite{li2025simplevla} & 28.1 & 37.5 & 77.3 & 45.3 & 47.1 \\
\midrule
\rowcolor{green!5}
\textbf{Ours} & \textbf{59.3} & \textbf{41.9} & \textbf{90.1} & 86.7 & \textbf{69.5} \\
\midrule

\multirow{2}{*}{\textbf{Model}} & \multicolumn{5}{c}{\textbf{Long and Extra Long Horizon Tasks (280--650 Steps)}} \\
\cmidrule(lr){2-6}
& \textbf{Handover Block} & \textbf{Stack Bowls Two} & \textbf{Blocks Rank Rgb} & \textbf{Put Bottles Dustbin} & \textbf{Avg.} \\
\midrule
$\pi_{0}$ \cite{black2024pi_0} & 39.0 & 53.0 & 45.0 & 54.0 & 47.8 \\
RDT \cite{liu2024rdt} & 26.0 & 42.0 & 17.0 & 26.0 & 27.8 \\
OpenVLA-OFT \cite{li2025simplevla} & 33.1 & 40.6 & 70.2 & 42.2 & 46.5 \\
\midrule
\rowcolor{green!5}
\textbf{Ours} & \textbf{50.4} & \textbf{69.9} & \textbf{77.2} & \textbf{57.8} & \textbf{63.8} \\
\bottomrule
\end{tabular}}
\end{table*}

\subsection{Results Across Benchmarks}
\label{sec:exp_result}

\textbf{Performance on LIBERO.}
\textbf{Table \ref{tab:libero_main}} presents the results on the LIBERO, showing that our method performs well across different tasks. Among them, we reach a higher success rate on LIBERO-Long, 
indicating that modeling planning actions can provide useful guidance for policy optimization.

\textbf{Performance on RoboTwin2.0.}
\textbf{Table \ref{tab:robotwin_main_horizon}} shows the performance on RoboTwin2.0 across short-horizon, medium-horizon, and long/extra-long-horizon task groups. Our method performs well on both short tasks and tasks that require longer and more structured execution. Additionally, PAPO-VLA shows advantages on long and extra-long tasks, indicating that planning actions modeling remain effective when needing to maintain correct behavior over extended action sequences. These results demonstrate the effectiveness of our method under diverse task settings and horizon lengths.

\subsection{Ablation Study}
\label{sec:abla}
\begin{wraptable}{r}{0.6\textwidth}
\vspace{-1.0em}
\centering
\caption{The effect of planning-action importance on LIBERO.}
\label{tab:abla}
\setlength{\tabcolsep}{3.2pt}
\renewcommand{\arraystretch}{0.95}
\small
\begin{tabular}{lccccc}
\toprule
\textbf{Model} & \textbf{Spatial} & \textbf{Object} & \textbf{Goal} & \textbf{Long} & \textbf{Avg.} \\
\midrule
Ours w/o Suff.\&Nec. & 0.85 & 0.88 & 0.79 & 0.53 & 0.76 \\
Ours w/o Suff. & 0.89 & 0.90 & 0.87 & 0.80 & 0.87 \\
Ours w/o Nec. & 0.88 & 0.92 & 0.89 & 0.85 & 0.89 \\
\textbf{Ours} & \textbf{0.93} & \textbf{0.98} & \textbf{0.98} & \textbf{0.94} & \textbf{0.96} \\
\bottomrule
\end{tabular}
\vspace{-1.0em}
\end{wraptable}

\textbf{The effect of different components.} We examine three ablated variants: (i) Ours w/o Causal Sufficiency \& Necessity (ii) Ours w/o Causal Sufficiency; (iii) Ours w/o Causal Necessity. The results in \textbf{Table \ref{tab:abla}} demonstrate the effectiveness of each component.

\textbf{Parameter sensitivity.}
We conduct experiments on the hyperparameter $\eta$ and search $\eta$ over $[0, 0.3]$. Results shows that the optimal result is at $\eta = 0.15$, which are our final configuration.

\subsection{Case Study and Visualization}
\label{sec:case_study}
To further illustrate the effectiveness, we visualize two representative tasks in \textbf{Figure \ref{fig:case}}: ``putting both moka pots on the stove'' and ``putting the yellow and white mug in the microwave and closing it''. Our method can produce more complete trajectories and maintain task progress over multiple stages.

\section{Conclusion}
\label{sec:conclusion}
This paper studies how to improve the reliability of VLA policies in robotic tasks. We revisit VLA execution and argue that a VLA policy plays two functional roles: a planner role that makes task-oriented decisions, and an executor role that realizes these decisions through dense continuous actions. Based on this view, we focus on planning actions, since they determine whether the following execution can remain on a successful path. We propose Planning-Aware Policy Optimization for VLA models (PAPO-VLA), which identifies planning actions from action variation and trajectory outcome, estimates their importance through causal sufficiency and causal necessity, and incorporates the resulting importance into GRPO advantage estimation. In this way, PAPO-VLA preserves trajectory-level optimization while assigning stronger emphasis to planning actions that are more important for task success. Extensive experiments demonstrate the effectiveness of PAPO-VLA.

\bibliographystyle{plainnat}
\bibliography{neurips_2026}

\end{document}